\documentclass{article}

\usepackage[final]{neurips_2023}

\usepackage[utf8]{inputenc} 
\usepackage[T1]{fontenc}    
\usepackage{hyperref}       
\usepackage{url}            
\usepackage{booktabs}       
\usepackage{amsfonts}       
\usepackage{nicefrac}       
\usepackage{microtype}      
\usepackage{xcolor}         

\usepackage{graphicx}
\usepackage{amsmath}
\usepackage{amssymb}
\usepackage{dsfont}
\usepackage{subcaption}
\usepackage{cleveref}
\usepackage{tabularx}
\usepackage{multirow}
\usepackage{pdfpages}
\usepackage{wrapfig}
\setcitestyle{square,numbers,comma}
\title{Harnessing Synthetic Datasets: The Role of Shape Bias in Deep Neural Network Generalization}

\author{%
  Elior Benarous
  \\
  ETH Zürich\\
  \texttt{ebenarous@ethz.ch}
  \And
  Sotiris Anagnostidis\\
  ETH Zürich
  \And
  Luca Biggio\\
  ETH Zürich
  \And
  Thomas Hofmann\\
  ETH Zürich
}

\newcommand{\viteight}{ViT-p8}
\newcommand{\vitsixt}{ViT-p16}

\begin{document}

\maketitle

\begin{abstract}
Recent advancements in deep learning have been primarily driven by the use of large models trained on increasingly vast datasets. While neural scaling laws have emerged to predict network performance given a specific level of computational resources, the growing demand for expansive datasets raises concerns. To address this, a new research direction has emerged, focusing on the creation of synthetic data as a substitute.
In this study, we investigate how neural networks exhibit shape bias during training on synthetic datasets, serving as an indicator of the synthetic data quality. Specifically, our findings indicate three key points: (1) Shape bias varies across network architectures and types of supervision, casting doubt on its reliability as a predictor for generalization and its ability to explain differences in model recognition compared to human capabilities. (2) Relying solely on shape bias to estimate generalization is unreliable, as it is entangled with diversity and naturalism. (3) We propose a novel interpretation of shape bias as a tool for estimating the diversity of samples within a dataset.
Our research aims to clarify the implications of using synthetic data and its associated shape bias in deep learning, addressing concerns regarding generalization and dataset quality.
\end{abstract}

\section{Introduction}
The success of deep learning models hinges on their ability to extract meaningful patterns from limited data, emphasizing the importance of comprehending the guiding principles of their learning process. In machine learning, inductive biases are foundational, representing the prior assumptions algorithms make about data structure. While neural networks are theoretically universal approximators \cite{HORNIK1989359}, practical constraints, such as finite training data and capacity, necessitate inductive bias to narrow the hypothesis space when encountering new instances \cite{HAUSSLER1988177}.

Inspired by advancements in neural network architecture, Vision Transformers (ViTs) have demonstrated superior performance in visual tasks compared to Convolutional Neural Networks (CNNs) \cite{og_vit}, provided sufficient computational resources. CNNs are designed around convolution and pooling operations, involving filters that capture local features, with inherent biases like translation equivariance and locality. Conversely, ViTs employ a self-attention mechanism, dividing input images into patches, linearly projecting them for embeddings, and subsequently passing these embeddings, along with positional encodings, through a Transformer encoder. This mechanism enables ViTs to capture global dependencies among patches, facilitating a more holistic understanding of image context and dependencies. This underlines the remarkable capabilities of ViTs in visual data.

In the pursuit of elucidating the exceptional performance of vision algorithms, a compelling avenue has emerged: comparing the inherent predilections of these algorithms with the proclivities of the human visual system. This exploration has yielded the concept of the "shape bias" as a pragmatic yardstick for quantifying disparities between machine learning models and human perceptual processes \cite{geirhos_main}. Notably, human cognition predominantly relies on shape-based recognition, constituting approximately \textasciitilde96\% of decision-making, while texture plays a minor role. Conversely, CNNs place substantial reliance on textural cues, with shape contributing to only about \textasciitilde25\% of their decision-making process.
In the realm of Vision Transformers, recent work by \citet{22bnparam} has spotlighted that with extensive training data -- surpassing 4 billion images in this instance -- ViTs not only attain state-of-the-art performance in out-of-distribution zero-shot classification and dense output tasks but also exhibit a remarkable shape bias, amounting to approximately \textasciitilde87\%. This achievement positions ViTs as the models closest to mirroring human perceptual biases. In light of these insights, we posit an intriguing question: \textit{could the shape bias exhibited by deep neural networks elucidate the gap between their recognition capabilities and the remarkable acumen of human vision?} To address this, we scrutinize the correlation between the shape bias and their generalization capacity.

In recent research, significant effort has been directed towards addressing challenges posed by limited training data. One effective approach is the use of data augmentations, as demonstrated in prior work \cite{NIPS2012_c399862d}. This technique proves especially valuable when dealing with smaller datasets that exhibit limited sample diversity. However, it is important to acknowledge the limitations of pixel-space augmentations. While they do help, they still maintain natural image statistics, which may lead models to rely on superficial cues \cite{DBLP:journals/corr/abs-1711-11561}.
Alternatively, a promising strategy involves leveraging synthetic datasets for pre-training purposes. By employing a trained generator, researchers can generate datasets of virtually unlimited size. This approach mitigates issues related to data bias and under-representation, offering particular advantages in scenarios with limited data availability, such as certain medical applications. Moreover, generative models offer a cost-effective alternative to the labor-intensive processes of data collection and annotation. They also sidestep potential privacy and legal concerns, as exemplified by the restrictions placed on the widely used ImageNet dataset due to privacy, ethics, and labeling issues \cite{pmlr-v81-buolamwini18a, Yang_2020}.

Recent studies (e.g., \citet{noise1} and \citet{https://doi.org/10.48550/arxiv.2101.08515}) have recently shown that utilizing unrealistic-looking synthetic samples for pre-training neural networks can significantly reduce the need for extensive datasets. In fact, some of these models outperform traditional ImageNet pre-trained models. According to \citet{noise1}, the key factor for effective training data is diversity and naturalistic characteristics, rather than strict realism. In other words, synthetic samples need not replicate real-world images precisely; they should instead capture essential structural properties of real data. Nevertheless, there is still no consensus on the specific properties that make a synthetic dataset most effective in harnessing its potential benefits.

This paper explores the role of shape bias in assessing a model's generalization potential on a synthetic dataset. We propose that dataset diversity is a critical factor in the dataset's quality, as it influences the model's shape bias across training epochs.
Our primary focus is to monitor the evolution of shape bias throughout training. Our key finding (1) challenges the conventional use of the shape bias metric as a reliable predictor for generalization, particularly when considering various architectural types and levels of supervision.
Furthermore, (2) we reveal the intricate relationships between shape bias, dataset diversity, and naturalism in the context of generalization. As a novel contribution, (3) we demonstrate the potential of shape bias as a reliable tool for measuring the diversity of training samples, presenting an alternative to methods like Fréchet Inception Distance (FID) \cite{fid}.

\section{Related Work}
\paragraph{Inductive bias.}

Inductive biases, whether explicit or implicit, are crucial constraints imposed by a model on the learning process, shaping its knowledge acquisition. These biases are paramount in determining what the model learns and how it learns.

Notably, specific biases emerge from the model's architecture \cite{https://doi.org/10.48550/arxiv.2211.14699}. Convolutional Neural Networks, known for their translation invariance, introduce a strong locality bias \cite{DBLP:journals/corr/abs-1906-06766}, even when initialized randomly. This bias enables CNNs to efficiently locate salient objects within images \cite{DBLP:journals/corr/abs-2106-09259}. Furthermore, the prevalent adoption of square pooling geometries amplifies a bias in favor of natural images \cite{DBLP:journals/corr/CohenS16a}, where adjacent pixels exhibit higher correlation. Adjusting these architectural choices can enhance CNNs' adaptability across various data types \cite{DBLP:journals/corr/CohenS16a}. Nevertheless, the locality bias inherent to CNNs hampers their ability to learn patterns extending over larger regions of the data.
Conversely, Vision Transformers leverage a self-attention mechanism that empowers them to capture both short and long-range dependencies between tokens, irrespective of their spatial separation. This stands in contrast to the limitations of CNNs, as discussed earlier.

The selection of training data introduces biases. Finite datasets exhibit inherent biases \cite{5995347}, prompting models to develop efficient strategies that excel within the training distribution (in-distribution \textit{i.d.}) but falter in out-of-distribution (\textit{o.o.d}) scenarios. Notably, CNNs trained on ImageNet have demonstrated a propensity for optimal performance with constrained receptive field sizes \cite{DBLP:journals/corr/abs-1904-00760} and resistance to global shape distortions while remaining sensitive to local shapes \cite{bioImagenet, 9522696}. This suggests that ImageNet-trained CNNs tend to emphasize textures over shapes \cite{geirhos_main}. Remarkably, CNNs can rapidly acquire a shape bias with minimal training data \cite{DBLP:journals/corr/abs-1802-02745}, leading to improved robustness against image distortions and higher \textit{o.o.d} accuracy \cite{DBLP:journals/corr/abs-1911-09071}. \citet{geirhos_main} and \citet{DBLP:journals/corr/abs-2010-05981} employ style transfer techniques \cite{7780634} to augment their training data with images that challenge the balance between shape and texture, ultimately yielding improved accuracy and adversarial robustness \citet{DBLP:journals/corr/abs-2010-05981}. This work underscores the complementary relationship between shape and texture biases in models.

\vspace{-1mm}
\paragraph{Learning from synthetic data.}
In recent developments, deep generative models, notably Generative Adversarial Networks (GANs), have emerged as valuable sources of synthetic data for improving classification \cite{https://doi.org/10.48550/arxiv.2106.05258, https://doi.org/10.48550/arxiv.1911.02888, https://doi.org/10.48550/arxiv.1905.10887} and semantic segmentation \cite{https://doi.org/10.48550/arxiv.2104.06490, https://doi.org/10.48550/arxiv.2103.04379, https://doi.org/10.48550/arxiv.2104.05833}. These models operate by learning from authentic image datasets and are adept at generating convincing synthetic samples.
In contrast, \citet{https://doi.org/10.48550/arxiv.2101.08515} have directed their focus towards fractals, intricate geometric patterns characterized by self-similarity across different scales, which can be produced through purely procedural methods. Such patterns are prevalent in the natural world, exemplified in structures like leaves, snowflakes, and seashells.
Furthermore, \citet{noise1} have undertaken an analysis of contrastive representation training using unrealistic data derived from statistical image models, procedural graphics models \cite{https://doi.org/10.48550/arxiv.2101.08515}, and samples drawn from generative models initialized at random. In our study, we place particular emphasis on the use of such unrealistic data samples, deliberately sidestepping potential biases and contamination issues frequently encountered in natural datasets.
\section{Method}
To comprehensively investigate the impact of shape bias, we evaluate both the ResNet architecture, representing CNNs, and the standard Vision Transformer architecture.
We conduct experiments with a ResNet-18 and a ResNet-50, as well as ViT-pX models, dividing the input into a grid of 8x8 and 16x16 patches, referred to as \viteight~and \vitsixt~respectively. Our customized ViTs all contain 7 encoder layers with 8 heads, and the hidden dimension is set to 512. 
\vspace{-1mm}
\paragraph{Training details.}
We train our models on 6 synthetic datasets.
First is the Fractals dataset, which we generate following the process provided by \citet{https://doi.org/10.48550/arxiv.2101.08515} (see Appendix \ref{apdx_train} for specifics).
The other 5 synthetic datasets are generated from a GAN.
Four are obtained through an untrained StyleGANv2  \cite{https://doi.org/10.48550/arxiv.1912.04958} whose parameters are sampled from different prior distributions.
These prior distributions are sufficient to encode image properties without requiring any training on real data.
A randomly initialized model is used to obtain StyleGAN - Random \ref{fig:viz_sg_rangom}. 
StyleGAN - High-freq. \ref{fig:viz_sg_highfreq} has supplementary noise maps and wavelets filters applied to increase the presence of noise. 
StyleGAN - Sparse \ref{fig:viz_sg_sparse} is intended to mimic the sparsity of natural images by adding a random bias to the convolutions and multiplying the noise maps to randomly sampled Laplacian noise.
StyleGAN - Oriented \ref{fig:viz_sg_oriented} further aims to approach the properties of real image by integrating oriented structures through tied wavelets.
The last dataset, Shaders-21k StyleGAN \ref{fig:viz_sg_shaders21k}, is obtained by sampling from a StyleGANv2, which was trained on the procedurally generated dataset Shaders-21k. 
The latter, itself, is created by applying 21 thousand generative procedures to shaders.
See Appendix~\ref{apdx_data_viz} to visualize samples from each class and \cite{https://doi.org/10.48550/arxiv.1912.04958, noise1, noise2} for further details on the creation of these synthetic datasets.
All datasets contain 95 thousand instances, sampled uniformly across generative parameters where applicable.
Each image is obtained in size 64x64, except for those in Shaders-21k StyleGAN, which we resize from 224x224 to 64x64.
Lastly, we test downstream \textit{o.o.d} performance on Tiny ImageNet.

For the synthetic datasets that do not contain labels, we train with self-supervision using the normalized temperature-scaled cross-entropy loss \cite{simclr}.
The latter maximizes the similarity between the representations of two augmented versions of a same image, while minimizing the similarity with all other samples in the batch.
The two versions are extracted as random crops of the same image.
We set the temperature to 0.5 and use the cosine similarity as a similarity measure.

Other training hyperparameters are kept constant across model architectures, datasets, and supervision types (see Appendix~\ref{apdx_train} for specifics).
This helps ensure a fair comparison between the aforementioned variables.
Prior to evaluation, we freeze the pre-trained encoder, randomly reset the fully connected layers, and fine-tune the latter with supervision during two epochs on the downstream dataset. 

Finally, the training samples undergo augmentation through different schemes, which vary between datasets.
For Tiny ImageNet, we apply the \textit{simCLR} augmentation scheme \cite{simclr}.
We transform the Fractals dataset with random horizontal flip, and random resized crop with a scale ranging between 8\% and 100\% of the original image size.
Finally, we apply the same augmentation scheme as \citet{noise1} to the GAN-generated samples.

\paragraph{Evaluation details.}
The benchmark dataset introduced by \citet{geirhos_main} to estimate the shape bias of a model is referred to as Cue conflict.
It is composed of 1200 images generated using style transfer between two ImageNet samples.
Consequently, these exhibit conflicting shape and texture features.
Each image is assigned two labels, indicating the shape and the transferred texture.
For each sample, we evaluate whether a model accurately classified the shape or texture.
To this end, we design an alternative method, distinct from that of \citet{geirhos_main}.
Rather than performing linear classification, we evaluate the embeddings with K-nearest-neighbours ($K$-NN) classification.
This addresses the need of employing a model with the same output classes as those of ImageNet, but also allows us to specifically analyze the encoder.
Given the smaller size of the dataset, we fix the number of neighbours to $K=5$.
We then classify the embedding of each sample twice, first according to the shape label of the remaining 1199 embeddings, and then according to their texture label.
In both cases, we use the cosine distance to determine the closest embeddings.
Ultimately, the shape bias is computed as the ratio between the number of samples correctly classified by their shape and the total number of samples correctly classified by either their shape or texture.
The bias hence ranges from 0 to 1, where 0 indicates a strong texture bias and 1 indicates a strong shape bias.

\section{Experiments and Results}
\subsection{Limitations of Shape Bias in the face of Architectural and Supervision Variations}
\begin{wrapfigure}{l}{0.5\textwidth}
\centering
\includegraphics[width=0.5\textwidth]{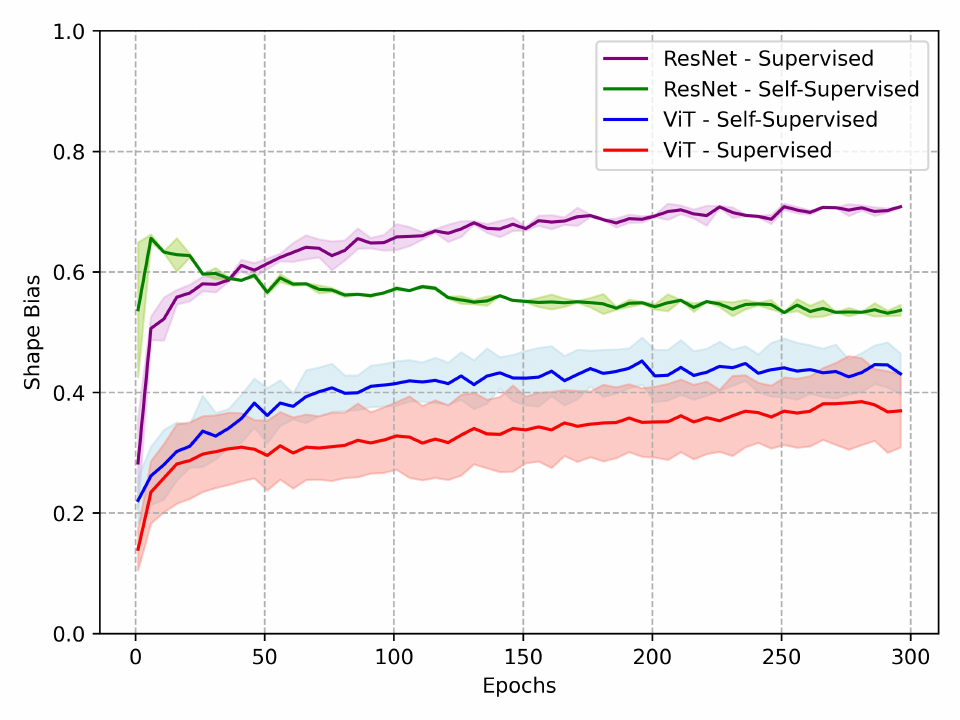}
\caption{Variation of the $K$-NN Shape Bias. We display the results averaged between ViT models (ViT-p8 and ViT-p16) and ResNet models (ResNet-18 and ResNet-50). Darker lines correspond to the averages and the associated filled regions have for upper and lower bounds the values of each model.}
\label{fig:tiny_one}
\vspace{-1mm}
\end{wrapfigure}

We first investigate the evolution of the shape bias during training on a realistic dataset, Tiny ImageNet, where the concept of shapes is more tangible.
Several observations emerge that shed light on the intricacies and nuances of this bias. 
These insights are pivotal in determining the validity of the shape bias metric as a reliable indicator of a dataset's capacity to teach models to generalize to real-world samples.

As depicted in Figure \ref{fig:tiny_one}, each combination of architecture and supervision follows a specific trend of its own.
All but the ResNets with self-supervision see their shape bias increase over the epochs.
Also, training with supervision induces a higher preference for shape with the convolution-based models, but not with the attention-based ones.

ResNets display the larger $K$-NN shape bias across supervision types.
However, this attempt to vaguely form groups according to architecture is limited by intra-class dissimilarities.
For instance, we notice a much larger shaded region for the ViTs, irrespective of the supervision type.
In both cases, the upper and lower bounds of the filled region correspond to the ViT-p8 and ViT-p16 respectively.
The sole distinction between these models is the number of patches they extract. 
This highlights a compelling correlation between the number of tokens and the resultant shape bias. 
This seems intuitive, as more patches equate to smaller segments of the image, thereby emphasizing local over global information.

These observations reveal that the shape bias is not solely a product of the training dataset.
It is induced quite distinctively by supervised and self-supervised learning, and is expressed differently by CNNs and ViTs.
This complexity challenges the feasibility of employing the shape bias as a robust and straightforward predictor for model generalization from synthetic to real-world datasets.
\subsection{Does Synthetic Data Generalize well because of Shape Bias?}

\begin{figure}[ht] 
    \centering
    \begin{subfigure}{0.49\textwidth}
        \centering
        \includegraphics[width=\textwidth]{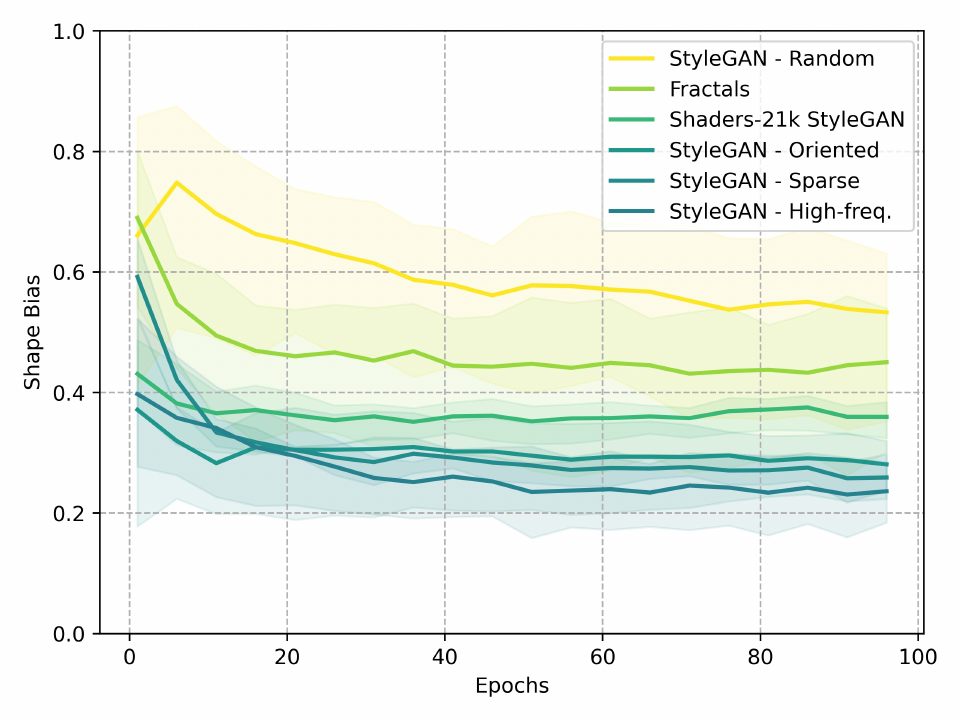}
        \caption{}
        \label{subfig:synthetic_bias_knn}
    \end{subfigure}
    \hfill
    \begin{subfigure}{0.49\textwidth}
        \centering
        \includegraphics[width=\textwidth]{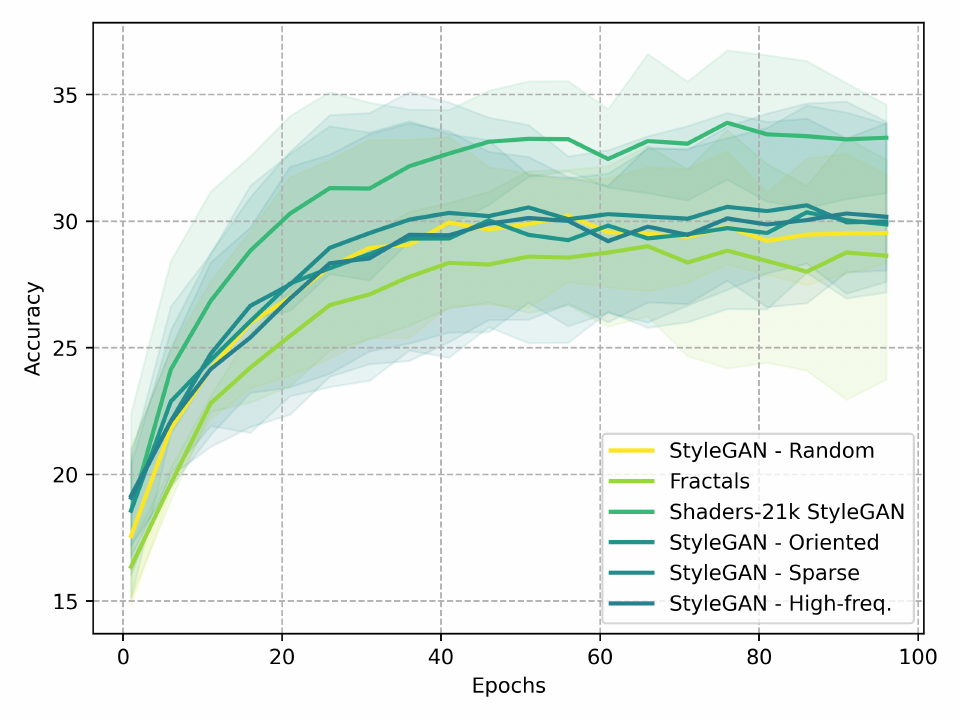}
        \caption{}
        \label{subfig:synthetic_acc_data}
    \end{subfigure}
\caption{Variation of the (a) $K$-NN Shape Bias and (b) Downstream accuracy on Tiny ImageNet's validation set over 100 epochs. In both plots, darker lines correspond to the average over the three architectures pre-trained on the same dataset. The associated shaded regions have for lower and upper bounds the values of the lowest and highest performing architectures respectively. For all datasets, the lower bound for the $K$-NN Shape Bias plot corresponds to the ViT-p8's performance. Colors follow the order of the average Shape Bias of each dataset after convergence (epoch 50 onward).}
\label{fig:synthetic_bias_acc}
\vspace{5mm}
\end{figure}

We now delve into the results obtained by training our models on the six synthetic datasets and testing them on Tiny ImageNet as our \textit{o.o.d} dataset.
For computational purposes, we focus on the ViT-p8 as our transformer architecture.
We track the performance of the models throughout 100 epochs of training.
We aim to determine whether there is a relation between the shape bias captured by a synthetic dataset and the generalization potential of a model pre-trained on that dataset.

An evident observation in Figure \ref{fig:synthetic_bias_acc} is that the shape biases generally start at a higher value compared to those seen with Tiny ImageNet training.
However, this value decreases over training epochs for all synthetic datasets.
One might think that this is only caused by the choice of self-supervision.
However this option is refuted by the evolution of the bias of the ViT-p8.
Indeed, the $K$-NN shape bias of this architecture kept increasing when trained on Tiny ImageNet with self-supervision.
However, it now also decreases from the first epoch onward.
Hence, some inherent, possibly shared, property of these datasets must be the cause for that decrease in shape bias.
As mentioned before, one plausible explanation, put forward by \citet{noise1}, is the need for diversity.

When comparing the datasets between one another, we find that StyleGAN - Random induces the highest shape bias among the six datasets.
Fractals and Shaders-21k StyleGAN take the second and third place respectively.
But most importantly, the last three are in a particularly interesting order.
Datasets that more closely resemble natural images achieved higher $K$-NN shape bias upon convergence.
This suggests that our evaluation of the shape bias is relevant to understand the properties of synthetic datasets, especially in the aim of having them mimic real-world images.
Nonetheless, one outlier remains, that is StyleGAN - Random.
Although being the farthest statistically from real images, this dataset induces the strongest shape bias.
Why is that?
As discussed above, this metric is influenced by many parameters.
Still, we can form the hypothesis that StyleGAN - Random samples are the ones that appear the smoothest visually, i.e. they contain the least texture.
In that sense, we can interpret it being unable to induce a strong texture bias to any model because its samples contain none (or very few) high frequency signals.
Consequently, even though a strong shape bias may be induced, this dataset is most likely lacking other crucial properties required in addition to shape bias for an overall strong generalization potential.
Such properties include those mentioned by \citet{noise1}: diversity and statistical similarity of samples with the downstream dataset.
When examining Figure \ref{subfig:synthetic_acc_data}, we find that this is indeed the case.
The order of the shape bias plot is lost and StyleGAN - Random does not stand out anymore on an \textit{o.o.d} recognition task.
Conversely, models pre-trained on Shaders-21k StyleGAN, supposedly the most diverse dataset, perform best.

Through our experiments on the six synthetic datasets, a key insight surfaced: while the $K$-NN evaluation of the shape bias offers valuable perspectives, it is not an all-encompassing metric for predicting generalization capability.
Intricately interwoven with shape bias are the fundamental elements of diversity and naturalism. 
A case in point are models pre-trained on StyleGAN - Random, which exhibit a pronounced shape bias. However, this bias might be a manifestation of its samples' texture paucity, which potentially compromises its diversity and naturalism. 
As a result, despite its strong shape bias, this dataset doesn’t surpass others in generalization.
\subsection{Evaluating the Diversity of a Dataset through the Shape Bias}
\begin{wrapfigure}{l}{0.5\textwidth}
\centering
\includegraphics[width=0.5\textwidth]{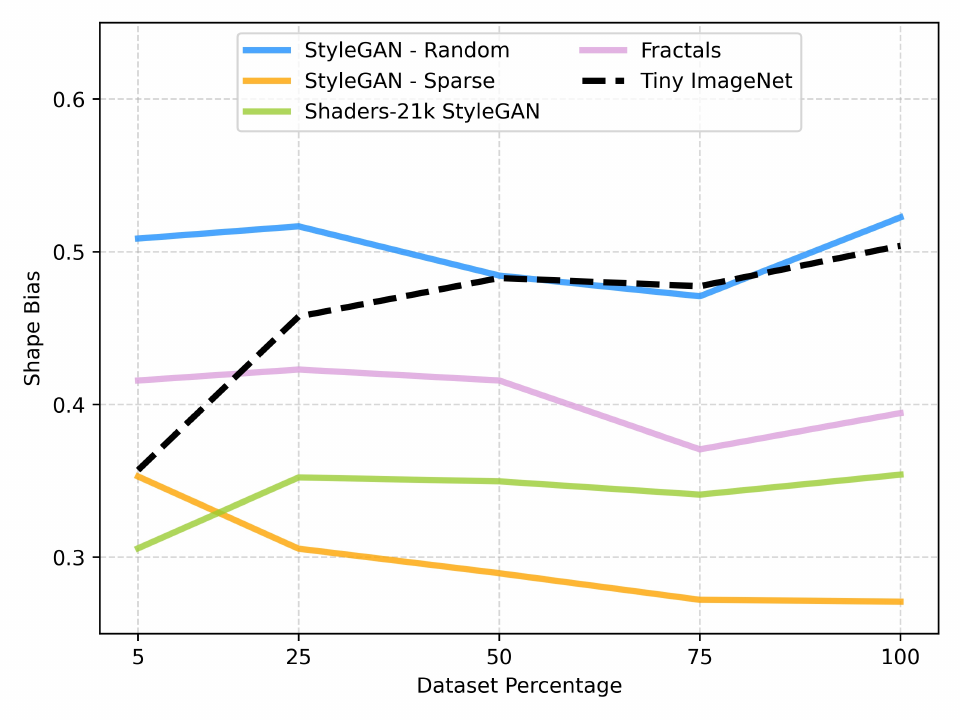}
\caption{Average value of the $K$-NN Shape Bias after convergence (epoch 50 onward). Results are averaged between the architectures we experimented with.}
\label{fig:scaling_bias_acc}
\end{wrapfigure}

Building on our earlier findings, we investigate the potential of shape bias as an estimator for dataset diversity.
The commonly adopted FID \cite{fid} score computes a distance using a pre-trained model, inevitably infusing the bias inherent to its training dataset. 
In contrast, our proposed method leverages shape bias for diversity estimation, eliminating the need for an auxiliary model trained on a distinct dataset.

In accordance with Occam’s razor, DNNs have the tendency to prioritize learning simple patterns of low frequency modes that generalize well \cite{DBLP:journals/corr/abs-1912-01198}, which \citet{pmlr-v97-rahaman19a} refer to as spectral bias.
Hence, we make the hypothesis that for highly diverse datasets, an increase in the number of samples would result in an increased shape bias. 
This is because a model with more data would be less prone to overfitting and, therefore, less likely to rely on high-frequency textures over low-frequency shapes. 
For this study, we pre-train models using varying amounts of synthetic data, specifically: 5\%, 25\%, 50\%, 75\% and 100\% of the total samples.
We focus on the ViT-p8 and ResNet-18 architectures for computational purposes.
Also, we center the discussion on StyleGAN - Random, due to its high $K$-NN shape bias in Cue conflict; Shaders-21k StyleGAN and Fractals, as both of which were visually very different from other datasets in our study; and StyleGAN - Sparse, because of its low shape bias but with some naturalistic elements by design.

The results displayed in Figure \ref{fig:scaling_bias_acc} align with our hypothesis.
We observe an increase of the shape bias with the number of Tiny ImageNet samples, which we consider being diverse.
In contrast, we notice very slim variations in the $K$-NN shape bias over the range of percentages for all synthetic datasets.
Models trained on the Fractals and StyleGAN - Sparse datasets reach a maximum shape bias at 25\% and 5\% of the initial number of samples respectively.
In a similar vein, the value at 25\% for Shaders-21k StyleGAN and StyleGAN - Random are lower than the peak value, reached with the complete dataset, by approximately 0.5\% and 1.1\% respectively.
In essence, the shape biases converge to consistent values, whether we use a reduced portion of the synthetic datasets or not.
These results suggest that the models keep relying on similar cues, mainly texture-based, regardless of the number of samples, which implies a deficiency in shape diversity within the datasets.

\section{Conclusion}
In this paper, we investigate the extent to which the shape bias of a model pre-trained on a synthetic dataset can be used to predict its generalization to realistic samples.
We first unveil the limitations of using such metric on any dataset due to the intricate dynamics with types of architecture and supervision.
We then highlight the entangled impact of shape bias, diversity and naturalism of synthetic samples on generalization.
Lastly, we suggest an alternative use of the shape bias as a proxy to estimate the diversity of samples.
We hope that this study spurs further research in achieving a holistic understanding of properties of synthetic datasets and in elucidating strategies for crafting effective synthetic datasets.


\clearpage
\bibliography{nips_bib}

\medskip

\clearpage
\appendix
\section*{Appendix}
\section{Further training details} \label{apdx_train}

\paragraph{Dataset generation.} 
Regarding the creation of the Fractals dataset, we follow the process defined by \citet{https://doi.org/10.48550/arxiv.2101.08515}.
We keep the intra-category augmentation parameters constant throughout.
With such, we sample 125 sets of fractal generation parameters and create 1000 images per set of parameters.
Accordingly, we obtain 125,000 samples.

\paragraph{Hyperparameters.}
We choose Adam \cite{kingma2017adam} as our optimizer for both pre-training and finetuning. 
A summary of the hyperparameters can be found in Table \ref{tab:adpx_param}.
These are kept identical for both pre-training and finetuning.

\begin{table}[h]
\vspace{5mm}
\centering
\begin{tabular}{ll}
\toprule
\textbf{Hyperparameter} & \textbf{Value} \\
\midrule
Adam $\beta_1$ & $0.9$ \\
Adam $\beta_2$ & $0.999$ \\
Learning Rate & $10^{-4}$ \\
Weight Decay & None \\
Batch Size & 64 \\
\bottomrule
\end{tabular}
\vspace{1mm}
\caption{Training hyperparameters and associated values.}
\label{tab:adpx_param}
\end{table}

\clearpage
\section{Dataset Visualization} \label{apdx_data_viz}
\subsubsection*{StyleGAN - Random}
\vspace{0.5cm}
\begin{figure}[h] 
    \begin{center}
    \includegraphics[scale=0.9]{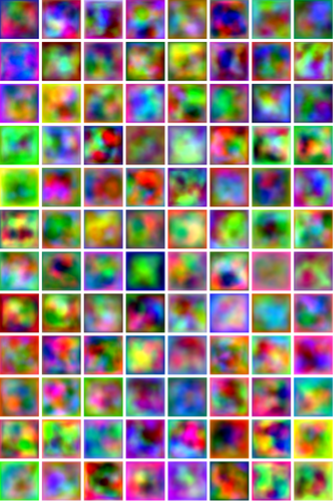} 
    \caption{96 random samples of the dataset StyleGAN - Random. Figure taken from \cite{noise1}}
    \label{fig:viz_sg_rangom}
    \end{center}
\end{figure}
\clearpage
\subsubsection*{StyleGAN - High freq.}
\vspace{0.5cm}
\begin{figure}[h] 
    \begin{center}
    \includegraphics[scale=0.9]{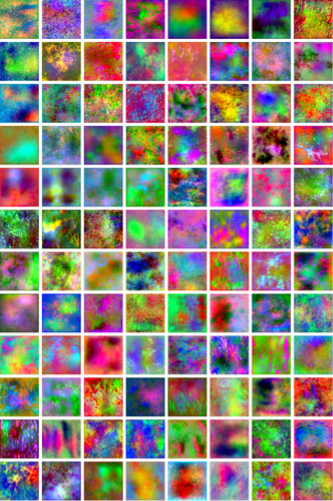} 
    \caption{96 random samples of the dataset StyleGAN - High freq.. Figure taken from \cite{noise1}}
    \label{fig:viz_sg_highfreq}
    \end{center}
\end{figure}
\clearpage
\subsubsection*{StyleGAN - Sparse}
\vspace{0.5cm}
\begin{figure}[h] 
    \begin{center}
    \includegraphics[scale=0.9]{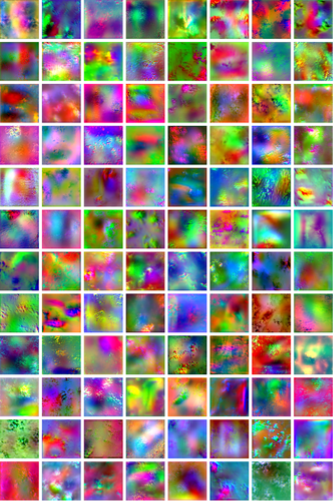} 
    \caption{96 random samples of the dataset StyleGAN - Sparse. Figure taken from \cite{noise1}}
    \label{fig:viz_sg_sparse}
    \end{center}
\end{figure}
\clearpage
\subsubsection*{StyleGAN - Oriented}
\vspace{0.5cm}
\begin{figure}[h] 
    \begin{center}
    \includegraphics[scale=0.9]{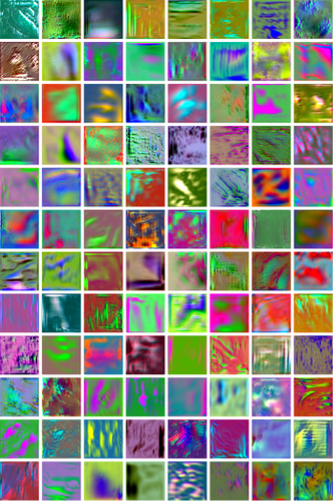} 
    \caption{96 random samples of the dataset StyleGAN - Oriented. Figure taken from \cite{noise1}}
    \label{fig:viz_sg_oriented}
    \end{center}
\end{figure}
\clearpage
\subsubsection*{S-21k StyleGAN}
\vspace{0.5cm}
\begin{figure}[h] 
    \begin{center}
    \includegraphics[scale=0.9]{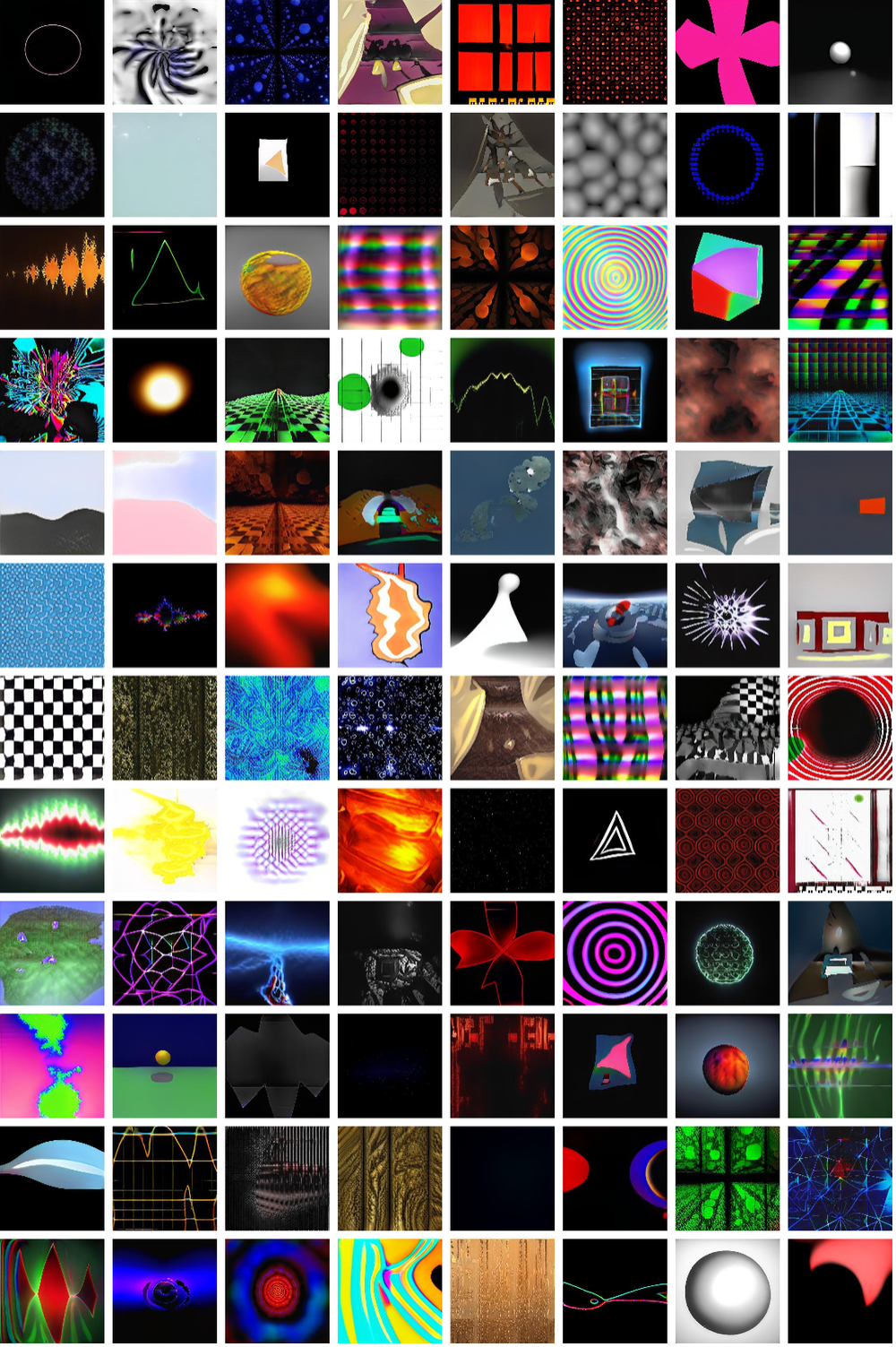} 
    \caption{96 random samples of the dataset S-21k StyleGAN. Figure taken from \cite{noise2}}
    \label{fig:viz_sg_shaders21k}
    \end{center}
\end{figure}
\clearpage
\subsubsection*{Fractals}
\vspace{0.5cm}
\begin{figure}[h] 
    \begin{center}
    \includegraphics[scale=1.05, clip, trim=2.4cm 11.6cm 5.7cm 2.4cm]{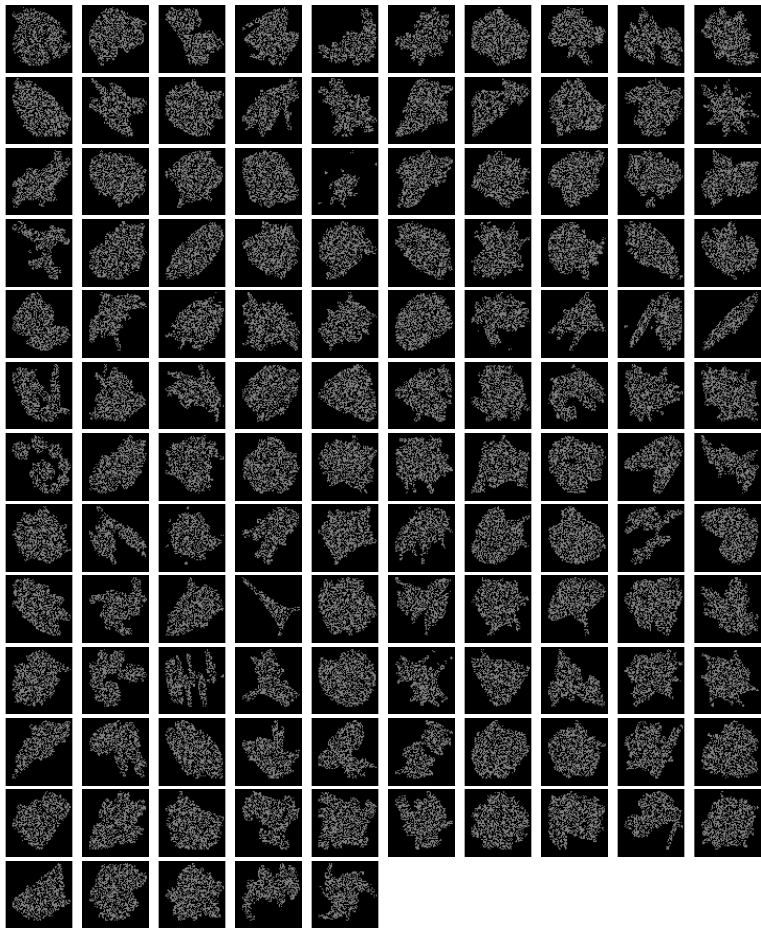} 
    \caption{125 random samples of the dataset Fractals}
    \label{fig:viz_fractal}
    \end{center}
\end{figure}

\end{document}